\documentclass[runningheads]{llncs}

\usepackage[T1]{fontenc}
\usepackage{amsmath}
\usepackage{amssymb}
\usepackage{booktabs}
\usepackage{graphicx}
\usepackage{url}
\usepackage[hidelinks,breaklinks=true]{hyperref}
\hypersetup{
  pdftitle={More Claims, Less Evidence: Bounded Verification of AI-Generated Digital Knowledge Artifacts},
  pdfauthor={Feliks Bańka and Jarosław A. Chudziak}
}

\graphicspath{{figures/}}

\begin{document}

\title{\texorpdfstring{More Claims, Less Evidence:\\Bounded Verification of AI-Generated\\Digital Knowledge Artifacts}{More Claims, Less Evidence: Bounded Verification of AI-Generated Digital Knowledge Artifacts}}
\titlerunning{More Claims, Less Evidence}

\author{Feliks Bańka \and Jarosław A. Chudziak}
\authorrunning{F. Bańka and J. A. Chudziak}
\institute{Institute of Computer Science, Warsaw University of Technology, Warsaw, Poland\\
\email{feliks.banka.stud@pw.edu.pl, jaroslaw.chudziak@pw.edu.pl}}

\maketitle

\begin{abstract}
Digital libraries, repositories, and AI-mediated knowledge services increasingly rely on generative systems to produce summaries, descriptions, and other multi-claim knowledge objects. Yet generation can scale far more easily than verification capacity: a reviewer may need to decide whether an object is suitable for publication or downstream use after checking only a small fraction of its claims. This creates a fundamental gap between claim-level verification and confidence in the object as a whole. The central question is therefore what a successful partial check implies about the reliability of the complete artifact when its size grows but the verification budget does not.
This paper contributes a Bayesian model of bounded verification centered on the \emph{Predictive Value of Pass} (PVP). The model predicts that evidentiary value decreases as artifacts grow under fixed verification capacity, improves with larger verification budgets, and is especially fragile when errors are sparse. Controlled experiments on FEVEROUS and FEVER support these predictions and show that adding supported claims around a fixed number of false or unsupported claims can make passing more likely while making a pass less informative.\footnote{Code and outputs: \url{https://github.com/latent-systems-lab/bounded-verification}}
The model further yields the minimum verification budget required to maintain a target PVP, providing a practical component for AI-assisted quality-assurance workflows in which generated knowledge objects must be checked before publication or downstream use.

\end{abstract}

\section{Introduction}
\label{sec:introduction}

Digital libraries, repositories, and other AI-supported knowledge services increasingly produce long-form outputs rather than isolated answers: summaries, reports, explanations, collection descriptions, and decision-support materials \cite{zerhoudi2024agents,yang2024dspace,mitchell2019modelcards,banka2026explanatoryequilibrium,banka2026multihedge}. These outputs can be broken down into statements about facts, evidence, assumptions, or intermediate reasoning steps~\cite{min2023factscore}. They support information access, review, and later use by people or other systems~\cite{gebru2021datasheets}. Reliability is assessed claim by claim, while the resulting knowledge object is typically consumed as a whole~\cite{wei2024safe}.

\begin{figure}[t]
\centering
\includegraphics[width=0.94\linewidth]{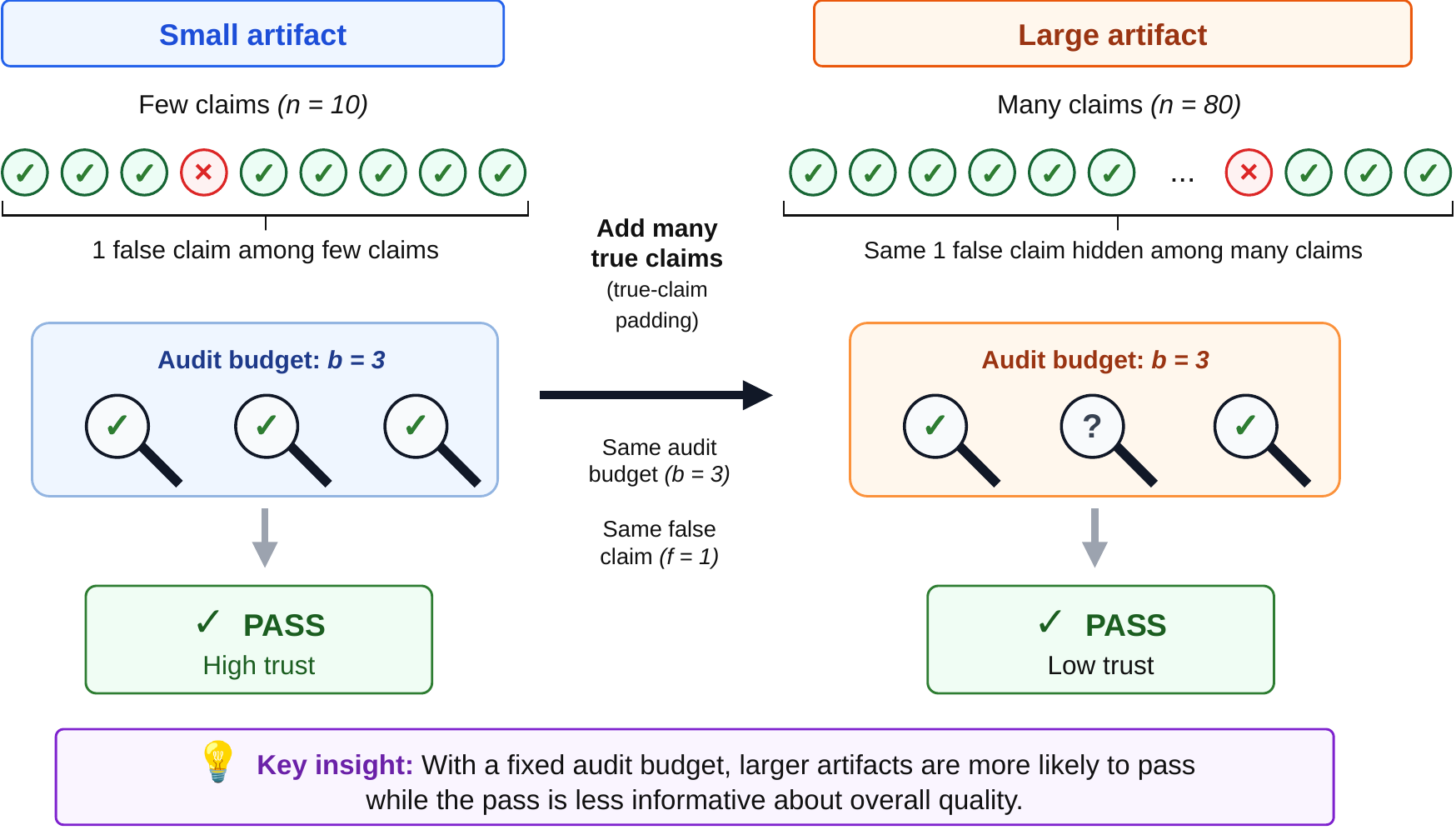}
\caption{Bounded verification of AI-generated digital knowledge artifacts. As artifact size grows under fixed verification capacity, unsupported claims can be diluted among supported ones, so a partial check provides posterior evidence, not a certificate of whole-object reliability.}
\label{fig:conceptual_overview}
\end{figure}

However, the ability to generate many claims does not imply the ability to verify all of them~\cite{min2023factscore,wei2024safe}. Classical sampling and inspection methods study settings in which only a limited subset of a larger population can be checked~\cite{dodge1947sampling,olives2013blqas,avenhaus2002inspectiongames}. For AI-generated knowledge objects, this creates a gap: generation can produce many plausible claims, while repository staff or other reviewers may be able to inspect only a small subset~\cite{mergel2026aiimplementation}. As illustrated in Figure~\ref{fig:conceptual_overview}, a small number of false or unsupported claims may therefore become diluted among many supported ones while the verification budget remains fixed. The key issue is not only whether a partial check detects an error, but how much trust should be placed in the complete object after it passes. In particular, adding supported claims around a fixed number of false or unsupported claims can make passing more likely while making a pass less informative. This matters in digital-library and repository settings because such a pass may influence whether a generated summary or description is shown to users or used by other services~\cite{yang2024dspace,xu2025subjectindexing}.

We study this problem using a Bayesian bounded-verification model. An object contains \(n\) auditable claims, a verifier checks only \(b\) of them, and an unreliable object may contain \(f\) false or unsupported claims. Our central quantity is the \emph{Predictive Value of Pass} (PVP), the posterior probability that an object is reliable given that it passed the bounded check. This shifts attention from error detection alone to the information conveyed by a successful check. We analyze how this value changes with object size, verification budget, and whether errors are sparse, and use the model in reverse to ask how much verification is needed to maintain a desired reliability level.

This paper provides a Bayesian interpretation of bounded verification for AI-generated digital knowledge objects. We define PVP for interpreting successful partial checks and show that, under fixed verification capacity, a pass becomes less informative as object size grows. We characterize the resulting true-claim padding effect, evaluate its dependence on object size, verification budget, and error sparsity using FEVEROUS and FEVER, and derive the minimum verification budget required to maintain a target PVP.

\section{Related Work and Positioning}
\label{sec:related}

Recent digital-library research applies LLMs to AI-supported access and conversational repository services~\cite{zerhoudi2024agents,yang2024dspace}. Related work uses language models for knowledge organization, subject indexing, and the evaluation of derived information structures~\cite{nandini2025organising,tan2025topic,xu2025subjectindexing}. Work on library AI implementation emphasizes human oversight, output quality, and resource constraints~\cite{mergel2026aiimplementation}, while interface interventions can encourage fact-checking of generated content~\cite{osop2025nudgr}. Related work on verifiable AI reasoning similarly uses explicit intermediate representations to create inspectable points within otherwise complex reasoning pipelines~\cite{Sadowski2025}. As these systems create or derive information objects for retrieval, description, and organization, their use raises questions about how much trust the complete object warrants when only some generated claims can be checked~\cite{yang2024dspace,xu2025subjectindexing}.

Acceptance sampling and sequential analysis make decisions from limited inspections rather than full population checks~\cite{dodge1947sampling,wald1947sequential}. Bayesian lot quality assurance sampling further connects bounded samples with predictive reliability~\cite{olives2013blqas}. We instead analyze posterior reliability after a partial audit passes, building on this predictive interpretation of limited inspection~\cite{olives2013blqas}. Our focus is how that reliability changes with artifact size under a fixed inspection budget~\cite{dodge1947sampling,avenhaus2002inspectiongames}.

FEVER and FEVEROUS evaluate whether claims are supported by textual or structured evidence~\cite{thorne2018fever,aly2021feverous}. Related benchmarks extend evidence-based verification to scientific claims, contrastive evidence, tables, and semi-structured data~\cite{wadden2020scifact,schuster2021vitaminc,chen2020tabfact,gupta2020infotabs}. Long-form factuality methods decompose generated outputs into verifiable claims, but checking every claim remains costly~\cite{min2023factscore,song2024veriscore}. Recent work therefore seeks more efficient factuality-evaluation pipelines~\cite{wei2024safe,wan2025fastfact}. These methods estimate or improve factuality~\cite{min2023factscore,wei2024safe}, whereas we ask a complementary receiver-side question: when only \(b\) of \(n\) claims can be checked, how much evidence does a successful check provide about the complete object as \(n\) grows?

\section{Bounded Verification of AI-Generated Digital Knowledge Artifacts}
\label{sec:model}

We now formalize bounded verification as a Bayesian inference problem. The model is deliberately minimal: an artifact consists of auditable claims, the verifier checks only a bounded subset, and the receiver updates trust after observing whether the artifact passes. We use uniform random sampling as a controlled baseline for isolating the effect of bounded verification capacity.

\subsection{Audit Model and Predictive Value of Pass}
\label{subsec:audit_model}

The audit model abstracts a generated artifact as a collection of claims, some of which may be false or unsupported. It has latent type \(\theta\in\{G,B\}\), where \(G\) denotes a reliable object and \(B\) one containing false or unsupported claims. The receiver has prior \(P(\theta=G)=p\) and \(P(\theta=B)=1-p\).

A bounded audit checks \(b\) claims sampled uniformly without replacement, with \(1\leq b\leq n\). When \(b=n\), the audit becomes full verification; our main interest is the bounded regime \(b\ll n\). A type-\(B\) object contains \(f\) false or unsupported claims, with \(1\leq f\leq n\). We focus on sparse-error settings, where \(f\ll n\), since a small number of problematic claims may then be hidden inside a large output. For example, an object containing 40 claims, 2 of which are false, passes an audit of 5 claims whenever the verifier misses both errors.

Formally, the artifact passes whenever none of the audited claims is false or unsupported. Thus, the bad-pass probability is
\[
q(n,f;b)
=
P(pass\mid \theta=B,f)
=
\frac{\binom{n-f}{b}}{\binom{n}{b}}.
\]
A reliable artifact passes with probability \(P(pass\mid \theta=G)=1\). Because \(q(n,f;b)\) only describes how often unreliable artifacts evade detection, we focus on posterior reliability after a pass.

The central quantity in this paper is the \emph{Predictive Value of Pass}:
\[
\mathrm{PVP}(n,f;b)
=
P(\theta=G\mid pass).
\]
By Bayes' rule,
\[
\mathrm{PVP}(n,f;b)
=
\frac{p}{p+(1-p)q(n,f;b)}.
\]

PVP measures how much trust a receiver should place in an artifact after it passes a bounded audit. This differs from the pass rate itself. An artifact can become more likely to pass as it grows, while the fact that it passed becomes less informative about its reliability.

\subsection{Evidentiary Dilution}
\label{subsec:dilution}

As \(n\) grows while \(f\) and \(b\) remain fixed, each false or unsupported claim occupies a smaller fraction of the artifact. A fixed-size audit is therefore less likely to sample a problematic claim. This is the mechanism behind evidentiary dilution.

\begin{proposition}[Evidentiary dilution]
\label{prop:dilution}
For fixed audit budget \(b\) and fixed false-claim count \(f\), over valid configurations with \(n\geq \max\{b,f\}\), \(\mathrm{PVP}(n,f;b)\) weakly decreases as artifact size \(n\) increases.
\end{proposition}

If \(n-f<b\), then \(q(n,f;b)=0\). Once \(n-f\geq b\), we use the product form
\[
q(n,f;b)
=
\prod_{j=0}^{b-1}\frac{n-f-j}{n-j}.
\]
Every factor is nonnegative and increases with \(n\), so \(q(n,f;b)\) is nondecreasing in this region. Moving from \(n-f<b\), where \(q=0\), into this region also preserves monotonicity. Hence PVP weakly decreases, proving the proposition.

To connect this quantity to decisions, suppose accepting a reliable artifact yields value \(V>0\), while accepting an unreliable artifact incurs loss \(L>0\). The receiver accepts after a pass when
\[
EU(Accept\mid pass)
=
\mathrm{PVP}(n,f;b)V
-
(1-\mathrm{PVP}(n,f;b))L
\geq 0.
\]
Equivalently, acceptance is justified only when \(\mathrm{PVP}(n,f;b)\geq \tau\), where \(\tau=L/(V+L)\). Thus, a successful audit may cease to justify acceptance once the artifact becomes sufficiently large relative to the audit budget.

For large \(n\), the bad-pass probability and the corresponding audit-budget rule are approximated by
\[
q(n,f;b)\approx e^{-bf/n},
\qquad
b\approx \frac{n}{f}\log\frac{1}{q_0}.
\]
Thus, maintaining a fixed bad-pass rate, and therefore a fixed PVP under a fixed prior, requires audit effort to grow roughly linearly with artifact size in sparse-error regimes.

\subsection{Required Verification Budget}
\label{subsec:required_budget}

The practical inverse question is how large the verification budget must be to maintain target posterior reliability \(\eta\). We therefore define
\[
b_{\min}(n,f,\eta)=\min\{b\in\{1,\ldots,n\}:\mathrm{PVP}(n,f;b)\geq\eta\}.
\]
The quantity $b_{\min}$ captures the verification effort required to maintain that reliability standard.
For \(p=0.5\) and \(f=2\), maintaining \(\mathrm{PVP}\geq0.8\) requires checking 5 of 10 claims but 40 of 80; for \(\mathrm{PVP}\geq0.9\), the corresponding budgets are 7 and 53.
Consistent with the scaling relation above, maintaining a fixed \(\eta\) under sparse errors requires verification effort to grow with \(n\).

\section{Empirical Evaluation of Model Predictions}
\label{sec:experiments}

We evaluate the model by applying bounded audits to artifacts constructed from labeled claims. We use gold labels as an oracle verifier to isolate verification capacity from verifier error.

\subsection{Research Hypotheses}
\label{subsec:hypotheses}

The model yields three testable predictions. \textbf{H1:} With fixed verification budget \(b\) and a fixed number \(f\) of false or unsupported claims, increasing artifact size \(n\) raises the bad-pass probability and lowers PVP. \textbf{H2:} Increasing \(b\) makes unsupported claims less likely to be missed and raises PVP. \textbf{H3:} Evidentiary dilution is stronger when errors remain sparse than when their number grows proportionally with \(n\).

\subsection{Experimental Setup}
\label{subsec:experimental_setup}

Benchmark-derived claims retain their natural-language wording and evidence while their support status is externally labeled. This allows us to vary object composition and audit capacity without conflating bounded verification with verifier accuracy. The evaluation is therefore a receiver-side stress test rather than an end-to-end factuality benchmark.

We use FEVEROUS as the main benchmark because it provides claim labels grounded in both textual and tabular evidence. We use FEVER as a text-only robustness check for the main true-claim padding result. In both datasets, supported claims are treated as reliable, while refuted or not-enough-information claims are treated as false or unsupported.

We construct each evaluated object as a set of benchmark claims. Reliable objects contain only supported claims, whereas unreliable ones contain \(n-f\) supported and \(f\) false or unsupported claims. Claims are sampled from dataset label pools without duplication within an object; reuse across objects is allowed. Unless otherwise stated, we use sizes \(n\in\{10,20,40,80\}\), false-claim counts \(f\in\{1,2,5\}\), audit budgets \(b\in\{3,5,10\}\), and prior \(p=0.5\). For each configuration and latent type, we construct \(1{,}000\) objects and run \(20\) independent audit simulations per object using a fixed random seed.

For each object, the audit samples \(b\) claims uniformly without replacement and passes if all are supported. We estimate the empirical bad-pass rate \(\widehat q(n)=\widehat P(pass\mid \theta=B)\), the overall pass rate, and the empirical Predictive Value of Pass \(\widehat{\mathrm{PVP}}(n)=\widehat P(\theta=G\mid pass)\). The overall pass rate is computed under the prior \(p=0.5\), so \(P(pass)=p+(1-p)\widehat q(n)\). The experimental configurations and parameter ranges are summarized in Table~\ref{tab:experimental_setup}. We compare these empirical quantities with the theoretical predictions from Section~\ref{sec:model}.

We report Wilson confidence intervals for empirical pass rates; they are small relative to the plotted effects and omitted from the main figures for readability.

\begin{table}[t]
\centering
\caption{Experimental configuration varying artifact size, audit budget, and number of false claims.}
\label{tab:experimental_setup}
\begin{tabular}{lll}
\toprule
Component & Values & Purpose \\
\midrule
Artifact size \(n\) & \(10,20,40,80\) & object growth \\
Audit budget \(b\) & \(3,5,10\) & verification capacity \\
False claims \(f\) & \(1,2,5\) & sparse errors \\
Main dataset & FEVEROUS & structured verification \\
Robustness & FEVER & text-only verification \\
\bottomrule
\end{tabular}
\end{table}

\subsection{Empirical Results}
\label{subsec:results_analysis}

The following experiments test the three model predictions by varying artifact size, verification budget, and error density.

\paragraph{E1: True-claim padding.}
E1 tests H1 by increasing artifact size through supported claims while holding \(b\) and \(f\) fixed.

\begin{figure}[t]
    \centering
    \includegraphics[width=\linewidth]{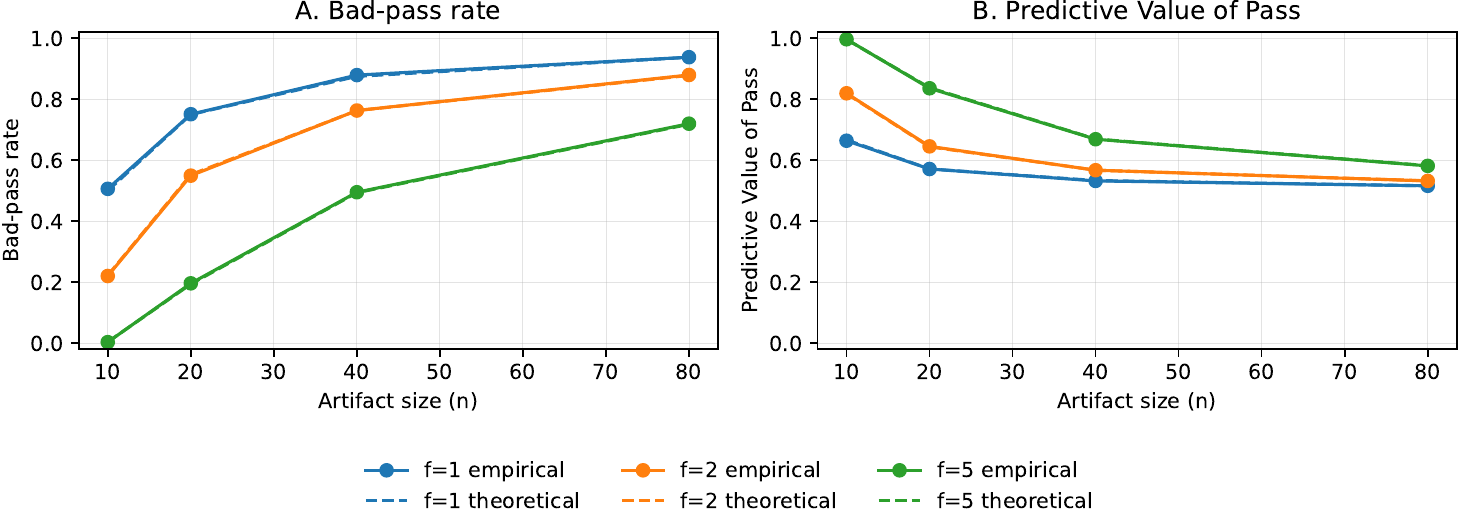}
    \caption{True-claim padding on FEVEROUS with audit budget \(b=5\): as artifact size grows, unreliable artifacts pass more often while PVP falls. Solid lines show empirical estimates and dashed lines the exact model.}
    \label{fig:true_claim_padding}
\end{figure}

Figure~\ref{fig:true_claim_padding} shows this pattern clearly. On FEVEROUS with \(b=5\) and \(f=2\), increasing artifact size from \(n=10\) to \(n=80\) raises the bad-pass rate from \(0.220\) to \(0.880\), while PVP falls from \(0.820\) to \(0.532\). The overall pass rate simultaneously increases from \(0.610\) to \(0.940\). Over the same range, verification coverage \(b/n\) falls from \(50\%\) to \(6.25\%\). Thus, artifacts pass more often, but passing becomes weaker evidence of reliability. The same qualitative trend appears for \(f=5\): the bad-pass rate increases from \(0.003\) at \(n=10\) to \(0.720\) at \(n=80\), while PVP decreases from \(0.997\) to \(0.581\).

Empirical estimates follow the exact model across configurations. The same pattern also appears on FEVER. For \(f=2\), increasing \(n\) from \(10\) to \(80\) raises the bad-pass rate from \(0.222\) to \(0.878\), while PVP decreases from \(0.818\) to \(0.533\).

\paragraph{E2: Audit budget sensitivity.}
E2 tests H2 by varying the audit budget \(b\). As predicted, larger audit budgets improve the evidentiary value of a pass by reducing the probability that all problematic claims are missed. For example, on FEVEROUS with \(n=80\) and \(f=5\), increasing \(b\) from \(3\) to \(10\) reduces the bad-pass rate from \(0.819\) to \(0.500\) and raises PVP from \(0.550\) to \(0.667\).

Increasing a fixed audit budget delays evidentiary dilution but does not remove the size effect as \(n\) continues to grow: larger budgets shift PVP upward, while each fixed-budget curve still declines with \(n\).

\paragraph{E3: Sparse versus denser errors.}
E3 tests H3 by comparing fixed sparse errors with errors that grow proportionally with \(n\). With sparse \(f=2\), PVP decreases from \(0.817\) at \(n=10\) to \(0.531\) at \(n=80\); when errors grow proportionally as \(f=\lceil0.20n\rceil\), PVP remains more stable, moving from \(0.819\) to \(0.761\). Large artifacts with a small number of hidden errors are therefore especially difficult to interpret under bounded auditing. This regime is relevant to otherwise high-quality generated knowledge objects, where most claims may be supported while a few consequential errors remain difficult to detect under bounded review.

\section{Discussion, Implications, and Limitations}
\label{sec:discussion}

A partial audit should not be treated as proof that the whole object is reliable: its value depends on object size, audit budget, and whether errors are sparse. The experiments highlight a pattern: bounded verification is least informative when an object is largely correct but contains only a few hidden errors, because those errors become easier to miss as the object grows. For generated summaries, descriptions, and briefs in repositories or digital-library services, quality assurance should either scale verification with the number of auditable claims or limit output length~\cite{mergel2026aiimplementation}. The required-budget calculation makes the needed verification effort explicit. In a repository workflow, a generated summary, description, or research brief can undergo bounded review~\cite{yang2024dspace}. PVP and the required-budget calculation can then indicate whether the result supports release, calls for more checks or fuller review, or warrants withholding before later use~\cite{xu2025subjectindexing}.

Our controlled study uses claim sets drawn from benchmarks with known binary labels, uniform random audits, and perfect verification of sampled claims. It separates the effect of verification capacity from verifier error but does not evaluate real objects generated by LLMs or human fact-checking~\cite{min2023factscore}, or automatic factuality verifiers~\cite{wei2024safe,wan2025fastfact}. Real objects may also contain claims of unequal importance and correlated errors.

Future work should measure these effects on repository objects under other audit policies. These include non-uniform inspection and imperfect human or automatic verification. It should also study natural LLM-generated repository or digital-library objects under these more realistic conditions.

\section{Conclusion}
\label{sec:conclusion}

AI-generated knowledge objects can grow faster than the resources available to verify them. This creates a gap between claim-level checking and trust in the whole object: when only part of an object can be inspected, a successful check does not by itself reveal how reliable the complete object is. We studied this question under bounded verification, where object size can grow while verification capacity remains limited.

Our main contribution is the Predictive Value of Pass (PVP), a Bayesian measure for interpreting successful partial verification. We show that adding supported claims around a fixed number of false or unsupported claims can make an unreliable object more likely to pass while making that pass less informative. Experiments on FEVEROUS and FEVER show the same effect, and the minimum-budget calculation shows how verification effort must scale to preserve a target reliability level. In AI-assisted curation workflows, PVP can therefore inform whether a generated object is ready for release after bounded review or requires additional verification.

\bibliographystyle{splncs04}
\bibliography{references}

\end{document}